\pdfoutput=1

\documentclass[11pt]{article}

\usepackage[preprint]{acl}

\usepackage{times}
\usepackage{latexsym}

\usepackage[T1]{fontenc}

\usepackage[utf8]{inputenc}

\usepackage{microtype}

\usepackage{inconsolata}

\usepackage{graphicx}

\usepackage{xspace}
\usepackage{amsmath}
\usepackage{amssymb}
\usepackage{xcolor}
\definecolor{Graylight}{gray}{0.9}
\definecolor{Gray}{gray}{1.0}
\usepackage{colortbl}
\usepackage{booktabs}
\usepackage{multirow}
\usepackage{mathrsfs}
\usepackage{marvosym}
\usepackage{enumerate}
\usepackage[hang,flushmargin]{footmisc} 
\usepackage{tablefootnote}
\usepackage{csquotes}

\newcommand{\huggingface}{\includegraphics[width=10px]{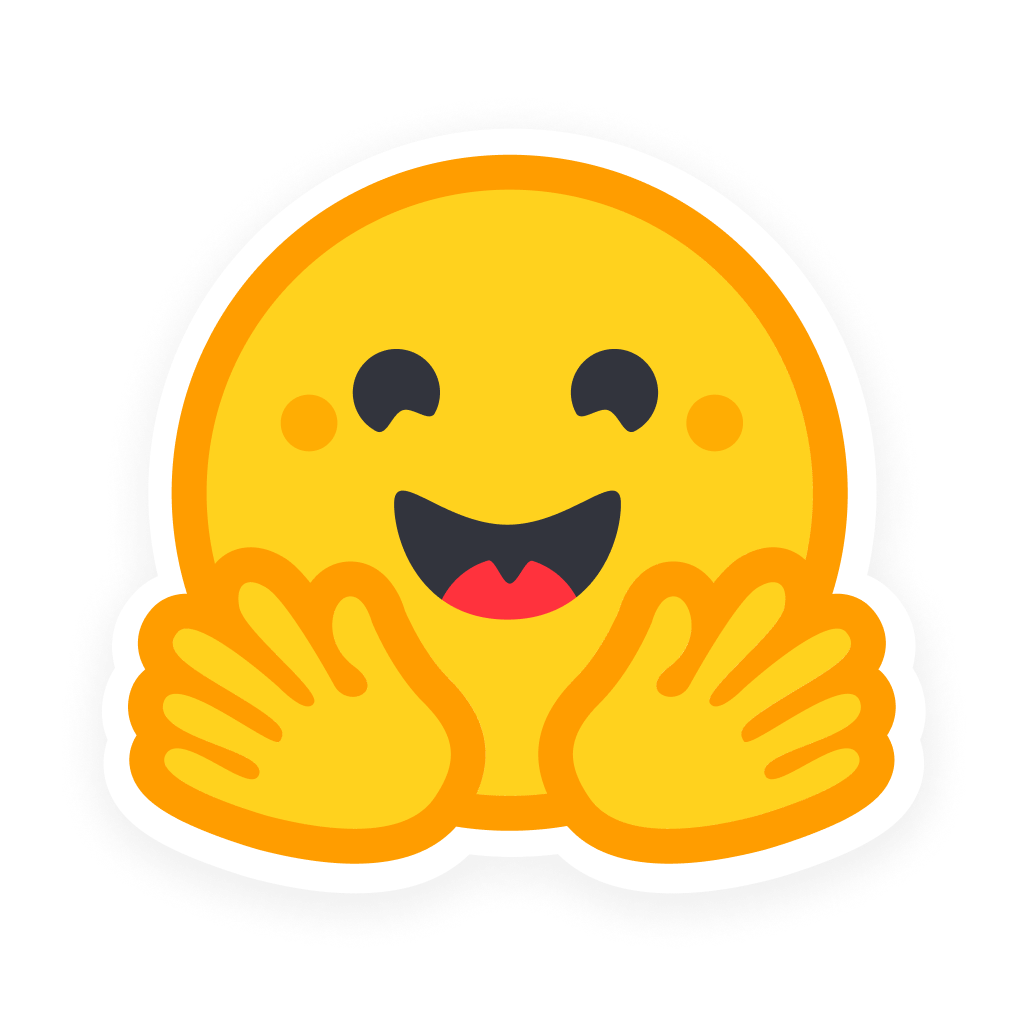}}

\newcommand{\sftname}{IXC-2.5\xspace}
\newcommand{\rewardname}{IXC-2.5-Reward\xspace}
\newcommand{\chatname}{IXC-2.5-Chat\xspace}

\title{InternLM-XComposer2.5-Reward: A Simple Yet Effective \\ Multi-Modal Reward Model}

\author{
  \textbf{Yuhang Zang\textsuperscript{*1}},
  \textbf{Xiaoyi Dong\textsuperscript{*1,2}},
  \textbf{Pan Zhang\textsuperscript{*1}},
  \textbf{Yuhang Cao\textsuperscript{*1}},
\\
  \textbf{Ziyu Liu\textsuperscript{$\dag$1,3}},
  \textbf{Shengyuan Ding\textsuperscript{$\dag$1,4}},
  \textbf{Shenxi Wu\textsuperscript{$\dag$1,5}},
  \textbf{Yubo Ma \textsuperscript{$\dag$1,6}},
\\
  \textbf{Haodong Duan\textsuperscript{1}},
  \textbf{Wenwei Zhang\textsuperscript{1}},
  \textbf{Kai Chen\textsuperscript{1}},
  \textbf{Dahua Lin\textsuperscript{1,2}},
  \textbf{Jiaqi Wang\textsuperscript{1}},
\\
  \textsuperscript{1}Shanghai Artificial Intelligence Laboratory,
  \textsuperscript{2}The Chinese University of Hong Kong,
\\
  \textsuperscript{3}Shanghai Jiao Tong University,
  \textsuperscript{4}Nanjing University,
\\
  \textsuperscript{5}Fudan University
  \textsuperscript{6}Nanyang Technological University
\\
openixclab@pjlab.org.cn
}

\begin{document}
\maketitle

{\let\thefootnote\relax\footnotetext{\noindent* indicates equal contribution. $\dag$ indicates interns at IXCLab, Shanghai AI Laboratory}}

\begin{abstract}
Despite the promising performance of Large Vision Language Models (LVLMs) in visual understanding, they occasionally generate incorrect outputs.
While reward models (RMs) with reinforcement learning or test-time scaling offer the potential for improving generation quality, a critical gap remains: publicly available multi-modal RMs for LVLMs are scarce, and the implementation details of proprietary models are often unclear.
We bridge this gap with InternLM-XComposer2.5-Reward (IXC-2.5-Reward), a simple yet effective multi-modal reward model that aligns LVLMs with human preferences.
To ensure the robustness and versatility of IXC-2.5-Reward, we set up a high-quality multi-modal preference corpus spanning text, image, and video inputs across diverse domains, such as instruction following, general understanding, text-rich documents, mathematical reasoning, and video understanding.
IXC-2.5-Reward achieves excellent results on the latest multi-modal reward model benchmark and shows competitive performance on text-only reward model benchmarks.
We further demonstrate three key applications of IXC-2.5-Reward: (1) Providing a supervisory signal for RL training. We integrate IXC-2.5-Reward with Proximal Policy Optimization (PPO) yields IXC-2.5-Chat, which shows consistent improvements in instruction following and multi-modal open-ended dialogue; (2) Selecting the best response from candidate responses for test-time scaling; and (3) Filtering outlier or noisy samples from existing image and video instruction tuning training data.
\end{abstract}    
\begin{figure*}
\centering
\includegraphics[width=0.9\textwidth]{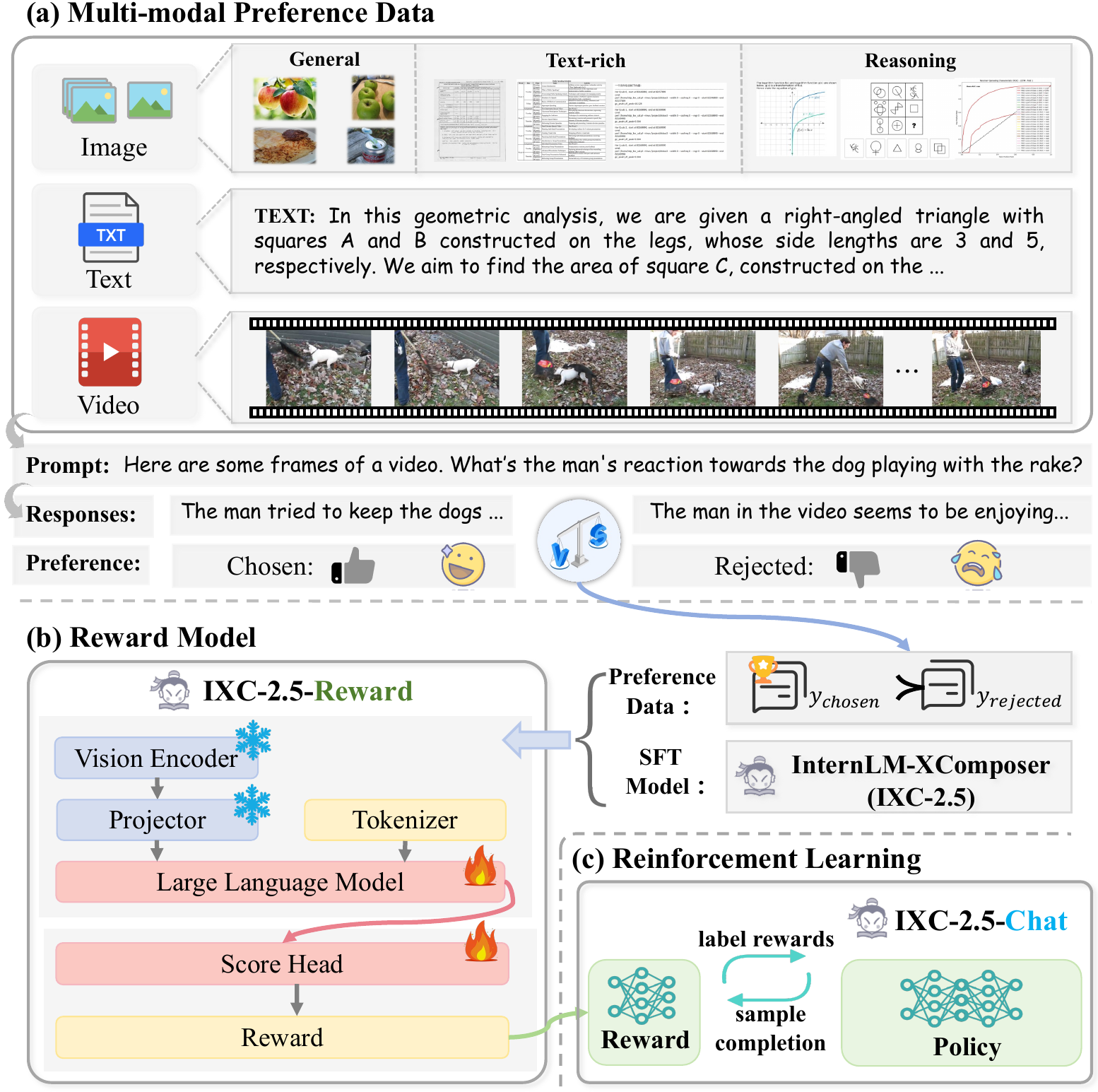}
\caption{\textbf{(a)} To train the \rewardname, we construct a multi-modal preference dataset spanning diverse domains (e.g., natural scenes, text-rich, reasoning) and modalities (image, text, video). \textbf{(b)} The framework of \rewardname. \textbf{(c)} The \rewardname guides policy training for \chatname via reinforcement learning.}
\label{fig:pipeline}
\vspace{-12pt}
\end{figure*}

\section{Introduction}
\leftline{``\emph{If you don't know where you are going, you'll end}}
\leftline{\emph{up some place else.}''}
\rightline{--- Yogi Berra}

Reward Models (RMs) \cite{cai2024internlm2,starling2023,liu2024skywork,wang2024helpsteer2,wang2024interpretable,yuan2024advancing,lou2024uncertainty,yang2024regularizing,yuan2024self,skyworkcritic2024,wang2024self} provide the crucial direction guidance about how well an AI model's outputs align with human preference, and benefit Large Language Models (LLMs) in training and inference.
During training, RMs are often used with reinforcement learning from human feedback (RLHF) \cite{ouyang2022training,bai2022constitutional,schulman2017proximal,rafailov2024direct} to penalize undesirable model behaviors and encourage outputs that align with human values.
At inference, RMs facilitate test-time scaling strategies \cite{snell2024scaling,gulcehre2023reinforced}, such as selecting the best response from candidate outputs or providing step-by-step critiques for complex reasoning tasks \cite{zelikman2022star,hosseini2024v}.


Despite their crucial role in both training and inference, multi-modal RMs for Large Vision Language Models (LVLMs) remain notably underexplored compared to language-only RMs for LLMs. Because current preference data is predominantly text-based and skewed toward specific domains (e.g., safety), data scarcity poses a significant challenge to training multi-modal RMs for diverse modalities such as images, videos, and text.
Consequently, existing multi-modal RMs \cite{wang2024rovrm,xiyao2024scaling} are largely constrained to narrow domains (e.g., mitigating hallucinations) or rely on prompting LVLMs with evaluation prompts, effectively functioning as generative RMs \cite{xiong2024llava}.
The limitation of multi-modal RMs subsequently constrains the capabilities of open-source LVLMs such as instruction following and safety-should-refuse, thereby hampering user interaction experience in multi-modal chat scenarios.

The growing community interest in RLHF and test-time scaling highlights the need for multi-modal RMs, which motivates us to present InternLM-XComposer2.5-Reward (\rewardname).
Instead of directly transferring unimodal (text) reward models (RMs) to the vision modality, we augment the existing LVLM (InternLM-XComposer2.5) with an additional scoring head to predict reward scores.
An effective multi-modal RM should ideally possess two key properties: (1) the ability to predict reward scores for both image, video, and textual inputs and (2) the capacity to generalize across diverse domains, such as instruction following, knowledge, text-rich images (e.g., documents), reasoning tasks, etc.
To this end, we develop a pipeline (Fig. \ref{fig:pipeline}(a)) to construct multi-modal preference data, and also incorporate existing high-quality datasets.
This pipeline selects prompts across diverse domains for text, image, and video inputs, generates corresponding responses, and then uses GPT-4o \cite{hurst2024gpt} or verifier \cite{lambert2024t} to perform preference judgments.
Trained on our preference data, \rewardname effectively evaluates both visual (image and video) and textual inputs (Fig. \ref{fig:pipeline} (b)).

\rewardname achieves best performance on multi-modal VL-RewardBench \cite{li2024vlrewardbench} (70.0\%) that beat all previous generative RMs including Gemini-1.5-Pro (62.5\%) and GPT-4o (62.4\%).
Even on uni-modal (text) RM benchmarks, \rewardname also demonstrates good results, with an average score of $88.6\%$ on Reward-Bench \cite{lambert2024rewardbench} and $68.8\%$ on RM-Bench \cite{liu2024rm}.

We further demonstrate the effectiveness of \rewardname in the following three aspects:

\noindent \textbf{(1)} \textbf{\rewardname for RL training.} We train a chat model (\chatname) using the on-policy Proximal Policy Optimization (PPO) algorithm with \rewardname to enhance its ability to follow instructions and provide a better user experience in multi-modal conversations.
Our results show clear improvements of \chatname on multi-modal instruction following and in-the-wild chatting benchmarks, which validate the effectiveness of \rewardname for providing the reward signal during RL training.

\noindent \textbf{(2)} \textbf{\rewardname for Test-Time Scaling.} Using best-of-$N$ sampling with \rewardname leads to additional performance gains compared to the RL-trained \chatname, confirming \rewardname's effectiveness in selecting good responses from candidate responses.

\noindent \textbf{(3)} \textbf{\rewardname for Data Cleaning.} We observe a strong correlation between low \rewardname scores and problematic samples, such as those exhibiting hallucinations or mismatched image/video and question/answer content. This suggests that \rewardname can effectively clean LVLM pre-training and post-training data.

\begin{table*}[t]
\centering
\begin{minipage}{0.43\textwidth}
\caption{Overview of existing preference datasets used in \rewardname. 
}
\vspace{-6pt}
\resizebox{.99\textwidth}{!}{
\begin{tabular}{c | c}
    \toprule
    \textbf{Category} & \textbf{Dataset} \\
    \midrule
    \multicolumn{2}{c}{\textbf{\textit{Text}}} \\
    \cmidrule(l{1em}r{1em}){1-2}
    \multirow{2}{*}{IF General} & Tulu-3-IF-augmented-on-policy-8b \cite{lambert2024t} \\
    ~ & UltraFeedback \cite{cui2024ultra} \\
    \cmidrule(l{1em}r{1em}){1-2}
    \multirow{2}{*}{Safety} & hhh alignment \cite{askell2021general}, PKU-Safe \cite{safe-rlhf} \\
    ~ & SHP \cite{ethayarajh2022understanding}, Anthropic-hhrlhf \cite{bai2022training} \\
    \midrule
    \multicolumn{2}{c}{\textbf{\textit{Image}}} \\
    \cmidrule(l{1em}r{1em}){1-2}
    Chat & WildVision-Battle \cite{lu2024wildvision} \\
    \cmidrule(l{1em}r{1em}){1-2}
    \multirow{2}{*}{General} & LLaVA-Critic \cite{xiong2024llava}, VL-Feedback \cite{li2024vlfeedback}, \\ ~ & RLAIF-V \cite{yu2024rlaifv}
    MIA-DPO \cite{liu2024miadpo} \\
    \bottomrule
    \end{tabular}}
    \vspace{-6pt}
\label{tab:data_used}
\end{minipage}
\hfill
\begin{minipage}{0.5\textwidth}
\centering
\caption{Overview of the source of newly collected data used in \rewardname.}
\vspace{-6pt}
\resizebox{.99\textwidth}{!}{
\begin{tabular}{c | c}
    \toprule
    \textbf{Category} & \textbf{Dataset} \\
    \midrule
    \multicolumn{2}{c}{\textbf{\textit{Image}}} \\
    \cmidrule(l{1em}r{1em}){1-2}
    \multirow{2}{*}{IF General} & MM-IFDPO-23k \cite{ding2025mm} \\
    ~ & KVQA \cite{shah2019kvqa}, A-OKVQA \cite{schwenk2022okvqa}, PMC-VQA \cite{zhang2023pmcvqa} \\
    \cmidrule(l{1em}r{1em}){1-2}
    \multirow{2}{*}{Text-Rich} & AI2D \cite{kembhavi2016diagram}, IconQA \cite{lu2021iconqa}, TQA \cite{kembhavi2017you} \\
    ~ & ChartQA \cite{masry2022chartqa}, DVQA \cite{kafle2018dvqa}, ScienceQA \cite{lu2022learn} \\
    \cmidrule(l{1em}r{1em}){1-2}
    \multirow{2}{*}{Reasoning} & GeoQA \cite{chen2021geoqa}, CLEVR-Math \cite{lindstrom2022clevr} \\
    ~ & Super-CLEVR \cite{li2023super}, TabMWP \cite{lu2022dynamic} \\
    \midrule
    \multicolumn{2}{c}{\textbf{\textit{Video}}} \\
    \cmidrule(l{1em}r{1em}){1-2}
    General & TrafficQA \cite{xu2021sutd}, FunQA \cite{xie2024funqa}, MiraData \cite{ju2024miradata} \\
    \bottomrule
    \end{tabular}}
\label{tab:data_new}
\vspace{-6pt}
\end{minipage}
\vspace{-6pt}
\end{table*}

\section{Related Work}

\noindent \textbf{Reward Model in Large Language Models.}
Reward models (RMs) are crucial for both Reinforcement Learning from Human Feedback (RLHF)~\cite{ouyang2022training,bai2022constitutional} and Test-time Scaling Laws~\cite{snell2024scaling,hosseini2024v}.
RMs have different implementation forms, such as
\textbf{(1)} discriminative RM \cite{cai2024internlm2,starling2023,liu2024skywork,wang2024helpsteer2,wang2024interpretable,yuan2024advancing,lou2024uncertainty,yang2024regularizing}, usually a sequence classifier that classifies input sequences into different categories, such as binary classification (``good'' or ``bad,'') or on a more granular scale \cite{wang2024helpsteer2,wang2024interpretable}. \textbf{(2)} generative RM~\cite{kim2023solar,yuan2024self,skyworkcritic2024,wang2024self} that are prompted to generate the feedback in the form of text, often a critique or explanation of why a certain output is good or bad.
\textbf{(3)} implicit RMs \cite{ivison2023camels,lambert2024t} that are models optimized using DPO \cite{rafailov2024direct} that the predicted log probabilities are interpreted as implicit reward signal.
Besides, RMs can also be divided into Outcome RMs (ORMs) \cite{cobbe2021training} and Process RMs (PRMs) \cite{uesato2022solving,lightman2023let,setlur2024rewarding}.
Our \rewardname belongs to the discriminative RM and ORM.

\noindent \textbf{Reward Model in Large Vision-Language Models.}
Previous RMs for LVLMs~\cite{wang2024rovrm,xiong2024llava,xiyao2024scaling} are limited to specific domains (e.g., reducing hallucination) or developed using relatively weak base models, which makes the implemented models significantly inferior to LLM RMs.
The lack of effective multi-modal RMs has created a bottleneck in vision RLHF, forcing researchers to merely use the variants of the off-poly DPO algorithm \cite{rafailov2024direct}. Previous work using open-source LVLMs as generative RMs \cite{yu2024rlaif,ouali2025clip,xiyao2024scaling}, injection of hallucinations with data augmentation techniques~\cite{deng2024enhancing,favero2024multi,zhou2024aligning,zhu2024self,pi2025strengthening,jiang2024modality,deng2024efficient} and rule-based selection~\cite{cao2024decompose,liu2024mia} for DPO data selection, which potentially compromise performance compared to the on-policy RL solutions like PPO \cite{schulman2017proximal}.
Moreover, lacking multi-modal RMs has also led to the reliance on human annotation \cite{sun2023aligning,yu2024rlhfv} or the use of proprietary models~\cite{zhang2024critic,zhao2024beyond} like GPT4 as generative RMs for DPO pair selection, which is expensive and unsustainable for large-scale applications.
Although open-source RMs for LVLMs have lagged behind their LLM counterparts, the growing community interest highlights the need for multi-modal RMs, which motivates our work.
In this work, we demonstrate that \rewardname is capable of combining with the PPO training and for DPO data selection at a low cost.

\noindent \textbf{Reward Model Evaluations.}
The development of evaluation benchmarks is essential for improving RMs.
Several comprehensive benchmarks have been proposed for evaluating RMs of LLMs, such as general abilities~\cite{lambert2024rewardbench,zhou2024rmb,liu2024rm}, multilingual~\cite{son2024mm,gureja2024m}, RAG \cite{yuan2024rag}, and mathematical process reward \cite{zheng2024processbench}.
The limited availability of multi-modal RMs has hampered the development of evaluation benchmarks, with existing benchmark~\cite{li2024vlrewardbench} focusing solely on generative RMs and lacking the evaluation of process supervision.
However, given the critical importance of RMs, we expect significant progress in multi-modal RM benchmarking in the future.

\section{IXC2.5-Reward}

\noindent \textbf{Data Preparation.} Reward models are trained using pairwise preference annotations (e.g., prompts $x$ with chosen responses $y_{c}$ and rejected responses $y_{r}$) that reflect human preferences.
While existing public preference data is primarily textual, with limited image and scarce video examples, we train \rewardname using both open-source data and a newly collected dataset to ensure broader domain coverage.

Tab. \ref{tab:data_used} lists the open-source pairwise data used in \rewardname, primarily focused on instruction following, safety, and general knowledge.
Tab. \ref{tab:data_new} details the source of our newly collected data, which is initially the supervised fine-tuning (SFT) data consisting of prompts $x$ and corresponding chosen responses $y_{c}$ across diverse domains: text-rich document understanding, math reasoning, and video understanding.
We also collect some in-house data about the instruction following, which will be released in the future.
To obtain rejected responses $y_{r}$, we prompt the SFT model, InternLM-XComposer-2.5 (\sftname) \cite{zhang2024ixc2d5} to generate multiple outputs for each prompt and then employ distinct selection criteria.
For general and text-rich data, we use GPT-4o \cite{hurst2024gpt} with pairwise evaluation prompts to determine the rejected response that was evaluated worse than the SFT ground-truth answer.
For math reasoning and instruction following data, we build verifier functions \cite{lambert2024t} that compare generated responses against ground-truth solutions to label the chosen and rejected data.
Our newly collected data complements existing open-source data, creating a comprehensive, high-quality multi-modal preference dataset.

\noindent \textbf{Model Architecture.} Our reward model InternLM-XComposer 2.5-Reward (\rewardname) is built upon the SFT model (\sftname) \cite{internlmxcomposer2_5}.
As shown in Fig. \ref{fig:pipeline} (b), we use the pre-trained weights of \chatname for most of the parts, such as the visual encoder and the MLP projector, which has aligned the image and video data with text modalities.
Thus, the \rewardname is merely required to train preference data to predict the reward score and avoid using other pre-training data for modality alignment.

We replace the final linear layer of \sftname with a score head $f$ for \rewardname that predicts the reward score.
Given the input prompt $x$ and the response $y$, the score head $f$ transforms the averaged hidden state features of all tokens into a binary scalar $r(x, y)$.
This scalar value $r(x, y)$ serves as the predicted reward score for the inputs.

\noindent \textbf{Loss Function.} Our reward model is trained via the following loss function:
\begin{equation}
    \mathcal{L}_{\text{RM}} = - E(\log(\sigma(r(x, y_{w}) - r(x, y_{l})))),
\end{equation}
where $r(x, y_{w})$ and $r(x, y_{l})$ denotes to the reward score assigned to the prompt $x$ with the chosen data $y_{w}$ and rejected data $y_{l}$, respectively.

\noindent \textbf{Training Strategy.} As shown in Fig.~\ref{fig:pipeline} (b), we froze the model's vision encoder and projector that are initialized from \sftname \cite{zhang2024ixc2d5}, training only the LLM (InternLM \cite{zhang2024ixc2d5}) and the score head. Other components of \sftname, such as the dynamic image partitioning mechanism for high-resolution inputs, remained unchanged.

\noindent \textbf{Length Constraints.} We remove data pairs where the length of the chosen response $y_{w}$ is significantly longer than the length of the rejected response $y_{l}$. This helps prevent the reward model from learning to associate length with quality.
Notably, we found that the vulnerability of LLM-based evaluation to length bias, a known issue in LLMs \cite{dubois2024length}, has also significant implications for LVLMs. Specifically, open-ended Visual Question Answering (VQA) benchmarks that employ LVLMs (e.g., GPT-4o) as judges are susceptible to inflated scores from overly long responses.
Consequently, removing the length constraint on the reward model resulted in improved PPO policy performance.
A detailed analysis is provided in Tab. \ref{tab:ablation_length}.

\begin{figure*}
    \centering
    \includegraphics[width=0.9\linewidth]{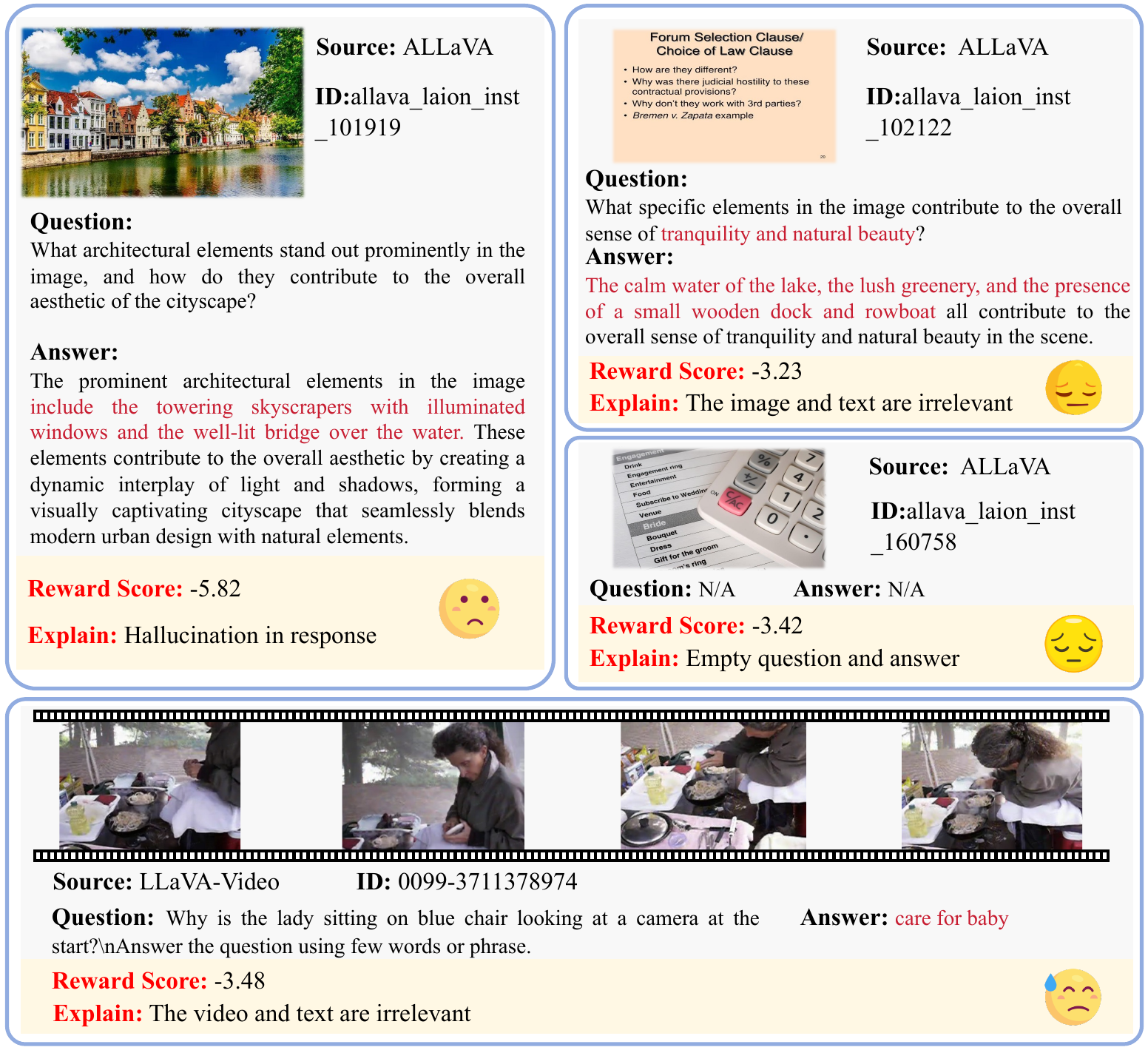}
    \vspace{-10pt}
    \caption{\textbf{Using \rewardname for Data Cleaning.} We visualize the outlier and noisy examples detected by \rewardname with low reward scores from existing image and video instruction-tuning datasets, such as ALLaVA \cite{chen2024allava} and LLaVA-Video-178K \cite{zhang2024video}. The “Explain” refers to explanations of error causes as identified by human experts, rather than outputs generated by the reward model.}
    \label{fig:data_clean}
    \vspace{-15pt}
\end{figure*}

\begin{table*}[t]
    \centering
    \small 
    \caption{\textbf{Evaluation results on VLRewardBench \cite{li2024vlrewardbench}.} The best and second-best results for proprietary models and open-source models are highlighted in \textbf{bold} and \underline{underlined}, respectively. }
    \vspace{-10pt}
    \resizebox{.95\textwidth}{!}{
    \begin{tabular}{@{}ll|ccc|cc@{}}
    \toprule 
    \textbf{Models}  & \textbf{\#Param}  &  \textbf{General}  & \textbf{Hallucination} & \textbf{Reasoning} & \textbf{Overall Acc}  & \textbf{Macro Acc} \\
    \midrule 
    \multicolumn{7}{c}{\emph{Proprietary Models}} \\ 
    \midrule 
    Gemini-1.5-Flash (2024-09-24)~\cite{geminiteam2024gemini15unlockingmultimodal} & - &	47.8&	59.6&	58.4& 	57.6 & 55.3\\ 
    Gemini-1.5-Pro (2024-09-24)~\cite{geminiteam2024gemini15unlockingmultimodal} & - & \textbf{50.8}&	\textbf{72.5}&	64.2& \textbf{67.2}	& \textbf{62.5}	\\ 
    
    Claude-3.5-Sonnet (2024-06-22)~\cite{anthropic2024claude} & - & 43.4&	55.0&	\underline{62.3} & 55.3 & 53.6\\ 
    GPT-4o-mini (2024-07-18)~\cite{gpt4o}   & - & 41.7&	34.5&58.2 & 41.5& 44.8\\ 
    GPT-4o (2024-08-06)~\cite{gpt4o}     & - & \underline{49.1} &	\underline{67.6} &	\textbf{70.5} & \underline{65.8} & \underline{62.4}\\ 
    \midrule
    \multicolumn{7}{c}{\emph{Open-Source Models}} \\ 
    \midrule
    LLaVA-OneVision-7B-ov~\cite{li2024llavaonevisioneasyvisualtask}  & 7B &	32.2 &	20.1  &	57.1  & 29.6 &36.5\\
    Qwen2-VL-7B~\cite{wang2024qwen2} & 7B &	31.6 &	19.1&	51.1 & 28.3 &33.9\\
    Molmo-7B~\cite{deitke2024molmopixmoopenweights}  & 7B	& 31.1& 	31.8& 	56.2 & 37.5 & 39.7\\
    InternVL2-8B~\cite{internvl2} &  8B	& 35.6&	41.1 &	59.0 & 44.5 &45.2\\
    LLaVA-Critic-8B~\cite{xiong2024llava} & 8B & \underline{54.6} & 38.3 & 59.1 & 41.2 & 44.0 \\
    Llama-3.2-11B~\cite{grattafiori2024llama3herdmodels} & 11B & 33.3 &	38.4 &	56.6 & 42.9 & 42.8\\ 
    Pixtral-12B~\cite{agrawal2024pixtral} & 12B & 35.6&	25.9 	& 59.9 & 35.8& 40.4\\
    Molmo-72B~\cite{deitke2024molmopixmoopenweights} & 72B	& 33.9&	42.3&	54.9 & 44.1  & 43.7\\
    Qwen2-VL-72B~\cite{wang2024qwen2} & 72B	& 38.1 &	32.8&	58.0 & 39.5 & 43.0\\ 
    NVLM-D-72B~\cite{dai2024nvlmopenfrontierclassmultimodal} &  72B & 38.9 &	31.6 & \underline{62.0} & 40.1 &  44.1\\
    Llama-3.2-90B~\cite{grattafiori2024llama3herdmodels} & 90B	& 42.6 & \underline{57.3} & 61.7 & \underline{56.2} & \underline{53.9}\\
   \cmidrule(r){1-2} \cmidrule(r){3-5} \cmidrule(r){6-7}
   \rowcolor[HTML]{F2F3F5}
   IXC-2.5-Reward (Ours) & 7B & \textbf{84.7} & \textbf{62.5} & \textbf{62.9} & \textbf{65.8} & \textbf{70.0} \\
    \bottomrule
    \end{tabular}}
    \label{tab:vl_reward}
    \vspace{-6pt}
\end{table*}

\section{The Applications of \rewardname}

In this section, we further validate three applications of \rewardname for (1) RL training (Sec. \ref{sec:rl}), (2) test-time scaling (Sec. \ref{sec:test-time}), and (3) data cleaning (Sec. \ref{sec:data-cleaning}).

\subsection{\rewardname for RL training}\label{sec:rl}
Having the reward model \rewardname enables the application of on-policy reinforcement learning algorithms (e.g., PPO \cite{schulman2017proximal}, RLOO \cite{ahmadian2024basic}, GRPO \cite{shao2024deepseek}) to optimize LVLM performance towards desired human preferences directly.
Using the PPO \cite{schulman2017proximal} algorithm, we subsequently train the policy model (\textbf{\chatname}, $\pi_{\theta}$) to maximize expected rewards from our reward model (\rewardname) while staying close to the reference model (\sftname, $\pi_{\text{ref}}$) for stability. A critic model $V$, initialized from \rewardname, is trained alongside $\pi_{\theta}$ to reduce the variance of policy updates.

\noindent \textbf{Data Prepration.} Similar to findings in \cite{hou2024chatglmrlhf}, we found that average reward scores differ across task domains (e.g., general, text-rich, reasoning).
This work focuses on improving the policy model's instruction following and open-ended chat abilities, which are critical for real-world applications such as stream chatting and human-AI interaction \cite{zhang2024internlm}.
Simultaneously, we ensure that performance in other domains (e.g., text-rich, reasoning) is not degraded relative to the SFT model \sftname.
Using our multi-modal preference data (which trains \rewardname), we curate a prompt set that prioritizes general chat and instruction following, while ensuring diversity through the inclusion of text-rich documents, math reasoning, and video understanding.

\noindent \textbf{PPO.} The PPO training begins by sampling a prompt from our prompt set. Then, the policy $\theta_{\pi}$ model generates responses, and the reward model computes the reward score $r_{t}$ at each state $s_{t}$ at the time-step $t$.
Given the reward score $r_{t}$ and and the critic model $V$, we compute the temporal difference error $\delta_{t}$, the Generalized Advantage Estimation (GAE) \cite{schulman2018high} $A_{t}$, and the Returns $R_{t}$ as:
\begin{equation}
\begin{aligned}
    \delta_{t} &= r_{t} + \gamma \cdot V(s_{t+1}) - V(s_{t}), \\
    A_{t} &= \delta_{t} + \gamma \cdot \beta \cdot A_{t+1}, \\
    R_{t} &= A_{t} + V(s_{t}),
\end{aligned}
\end{equation}
where $\gamma$ is a discount factor that determines how much future rewards are valued compared to immediate rewards, and $\beta$ is the parameter that controls the trade-off between bias and variance in the advantage estimation.
The advantage $A$ refers to how much better the policy model did than expected, and the returns $R$ is the cumulative reward.

Based on the advantage $A$, we compute the policy gradient loss $\mathcal{L}_{\text{PG}}$ to update the policy model $\pi_{\theta}$:
\begin{equation}
    \label{eq:pg}
    \mathcal{L}_{\text{PG}} = \min(\frac{\pi_{\theta}}{\pi_{\text{ref}}} \cdot A, \text{clip}(\frac{\pi_{\theta}}{\pi_{\text{ref}}}, 1.0 - \epsilon, 1.0 + \epsilon) \cdot A),
\end{equation}
where $\frac{\pi_{\theta}}{\pi_{\text{ref}}}$ is the log of the probability ratio between the policy model $\pi_{\theta}$ and the reference model $\pi_{\text{ref}}$, and $\epsilon$ is a hyper-parameter that controls the clipped ratio.

We further update the critic model via the Mean Squared Error (MSE) loss to minimize the difference between the predicted value of a state $V(s_{t})$ and the actual return $R_{t}$ obtained from state $t$:
\begin{equation}
    \mathcal{L}_{\text{critic}} = \sum_{t} \text{MSE}( V(s_{t}), R_{t} ).
\end{equation}

In summary, with the help of \rewardname and PPO, we train the \chatname to generate responses that improve the quality of multi-modal chat and follow user instructions.
The quality of \chatname also demonstrates the quality of \rewardname that provides the reward scores.

\subsection{\rewardname for Test-Time Scaling}\label{sec:test-time}

We further demonstrate that \rewardname is essential for scaling the inference-time capabilities of LVLMs.
We select the Best-of-N (BoN) sampling technique that improves the quality of generated text by using the reward model.
Specifically, the \chatname model generates multiple ($N$) different text outputs with different random seeds for a given prompt.
Subsequently, the reward model \rewardname scores each of these $N$ outputs, and the output with the highest score from the reward model is selected as the final output.

\subsection{\rewardname for Data Cleaning}\label{sec:data-cleaning}

Garbage in, garbage out.
Problematic samples in instruction tuning datasets negatively impact LVLM training.
While existing methods \cite{chen2024sharegpt4video} employ classifiers like CLIP \cite{radford2021learning} for filtering, these approaches have limitations, particularly with long-context inputs \cite{zhang2025long}, high-resolution images, or videos.
As shown in Fig. \ref{fig:data_clean}, we observe a strong correlation between low \rewardname scores and problematic samples, including hallucinations, empty answers, and irrelevant image/video-text pairings.
Therefore, \rewardname effectively cleans both pre-training and post-training data for LVLMs.
\section{Experiments}

\begin{table*}[t]
    \centering
    \small 
    \caption{\textbf{Evaluation results on Reward Bench \cite{lambert2024rewardbench}.} We report the performance of selective representative language-only RMs and previous multi-modal generative RMs.}
    \vspace{-6pt}
    \resizebox{.9\textwidth}{!}{
    \begin{tabular}{@{}ll| cccc|c @{}}
    \toprule
    Model Name & LLM & Chat & Chat Hard &  Safety & Reasoning & Avg Score \\
    \midrule
    \multicolumn{7}{c}{\emph{Language-Only Reward Models}} \\
    \midrule
    InternLM2-7B-Reward~\cite{cai2024internlm2} & InternLM2-7B & 99.2 & 69.5 & 87.2 & 94.5 & 87.6 \\
    InternLM2-20B-Reward~\cite{cai2024internlm2} & InternLM2-20B & 98.9 & 76.5 & 89.5 & 95.8 & 90.2 \\
    Skyword-Reward-Llama3.1-8B~\cite{liu2024skywork} & Llama3.1-8B & 95.8 & 87.3 & 90.8 & 96.2 & \underline{92.5} \\
    INF-ORM-Llama3.1-70B~\cite{informllama2024} & Llama3.1-70B & 96.6 & 91.0 & 93.6 & 99.1 & \textbf{95.1} \\
    \midrule
    \multicolumn{7}{c}{\emph{Multi-Modal Reward Models}} \\
    \midrule
    QWen2-VL-7B~\cite{wang2024qwen2} & QWen2-7B & 96.6 & 57.0 & 73.9 & 94.3 & 83.8 \\
    LLaVA-Critic-8B~\cite{xiong2024llava} & LLaMA3-7B & 96.9 & 52.8 & 81.7 & 83.5 & 80.0 \\
    \cmidrule(r){1-2} \cmidrule(r){3-6} \cmidrule(r){7-7}
    \rowcolor[HTML]{F2F3F5}
    IXC-2.5-Reward (Ours) & InternLM2-7B & 90.8 & 83.8 & 87.8 & 90.0 & 88.6 \\
    \bottomrule
    \end{tabular}}
    \label{tab:reward-bench}
    \vspace{-12pt}
\end{table*}

\begin{table*}[t]
    \centering
    \small 
    \caption{\textbf{Evaluation results on RM-Bench \cite{liu2024rm}.} We classify reward models into three types: sequence classifiers (Seq.), generative models, and implicit DPO models. Performance is reported across four domains (Chat, Math, Code, Safety) and three difficulty levels (Easy, Normal, Hard), along with average scores.}
    \vspace{-6pt}
    \resizebox{.95\textwidth}{!}{
    \begin{tabular}{@{}ll|cccc|ccc|c @{}}
    \toprule
    Model Name & Type & Chat & Math & Code & Safety & Easy & Normal & Hard & Avg \\
    \midrule
    \multicolumn{10}{c}{\emph{Language-Only Reward Models}} \\
    \midrule
    Tulu-2-dpo-13b~\cite{ivison2023camels} & Implicit & 66.4 & 51.4 & 51.8 & 85.4 & 86.9 & 66.7 & 37.7 & 63.8 \\
    InternLM2-7B-Reward~\cite{cai2024internlm2} & Seq. & 61.7 & 71.4 & 49.7 & 85.5 & 85.4 & 70.7 & 45.1 & 67.1 \\
    InternLM2-20B-Reward~\cite{cai2024internlm2} & Seq. & 63.1 & 66.8 & 56.7 & 86.5 & 82.6 & 71.6 & 50.7 & 68.3 \\
    Nemotron-4-340B-Reward~\cite{wang2024helpsteer2} & Generative & 71.2 & 59.8 & 59.4 & 87.5 & 81.0 & 71.4 & 56.1 & 69.5 \\
    URM-LLaMa-3.1-8B~\cite{lou2024uncertainty} & Seq. & 71.2 & 61.8 & 54.1 & 93.1 & 84.0 & 73.2 & 53.0 & \underline{70.0} \\
    Skyword-Reward-Llama3.1-8B~\cite{liu2024skywork} & Seq. & 69.5 & 60.6 & 54.5 & 95.7 & 89.0 & 74.7 & 46.6 & \textbf{70.1} \\
    \midrule
    \multicolumn{10}{c}{\emph{Multi-Modal Reward Models}} \\
    \midrule
    \rowcolor[HTML]{F2F3F5}
    IXC-2.5-Reward (Ours) & Seq. & 65.5 & 55.9 & 51.7 & 93.8 & 87.5 & 71.3 & 47.4 & 68.8 \\
    \bottomrule
    \end{tabular}}
    \label{tab:rm-bench}
    \vspace{-6pt}
\end{table*}

\begin{table*}[t!]
    \centering
    \small 
    \caption{Evaluation results of our IXC-2.5-Chat model against previous SOTA proprietary and open-source models $\le$10B (results are copied from \href{https://huggingface.co/spaces/opencompass/open_vlm_leaderboard}{\huggingface OpenVLM Leaderboard} and \href{https://huggingface.co/spaces/opencompass/Open_LMM_Reasoning_Leaderboard}{\huggingface Open LMM Reasoning Leaderboard}, accessed 01-Jan-2025).
    \textbf{Best} and \underline{second best} results are highlighted.}
    \vspace{-6pt}
    \resizebox{.95\textwidth}{!}{
    \begin{tabular}{@{}llc| c| ccc @{}}
    \toprule
    \multirow{2}{*}{Category} & \multirow{2}{*}{Benchmark} & \multirow{2}{*}{Evaluation} & Proprietary API & \multicolumn{3}{c}{Open-Source Model \textbf{($\le$10B)}} \\
    ~ & ~ & ~ & Previous-SOTA & Previous-SOTA & IXC-2.5 & IXC-2.5-Chat \\
    \midrule
    Instruction  & WildVision$_{(0617)}$ \cite{lu2024wildvision} & Open & 89.2 \cite{hurst2024gpt} & 67.3 \cite{xiong2024llava} & 37.5 & \cellcolor[HTML]{F2F3F5} 74.6 \\
    Following & MIA$_{(\text{val})}$ \cite{qian2024mia} & Open & 88.6 \cite{hurst2024gpt} & 80.7 \cite{wang2024qwen2} & 80.4 & \cellcolor[HTML]{F2F3F5} 84.0 \\
    \& Chat & MM-MT$_{(\text{val})}$ \cite{agrawal2024pixtral} & Open & 7.72 \cite{hurst2024gpt} & 5.45 \cite{wang2024qwen2} & 3.85 & \cellcolor[HTML]{F2F3F5} 5.70 \\
    ~ & MM-Vet v2$_{(\text{0613})}$ \cite{yu2023mm} & Open & 71.8 \cite{anthropic2024claude} & 58.1 \cite{chen2024expanding} & 45.8 & \cellcolor[HTML]{F2F3F5} 54.8 \\
    \midrule
    Knowledge & MMBench$_{(\text{v1.1})}$ \cite{liu2025mmbench} & MCQ & 85.7 \cite{sensetime2024sense} & 82.7 \cite{lu2024bluelm} & 79.4 & \cellcolor[HTML]{F2F3F5} 79.0 \\
    ~ & MMMU$_{(\text{val})}$ \cite{yue2024mmmu} & MCQ & 70.7 \cite{hurst2024gpt} & 56.2 \cite{chen2024expanding} & 42.9 & \cellcolor[HTML]{F2F3F5} 44.1 \\
    ~ & MMStar \cite{chen2024we} & MCQ & 72.7 \cite{sensetime2024sense} & 63.2 \cite{chen2024expanding} & 59.9 & \cellcolor[HTML]{F2F3F5} 59.6 \\
    \midrule
    Reasoning & MathVista$_{(\text{mini})}$ \cite{lu2023mathvista} & VQA & 78.4 \cite{sensetime2024sense} & 66.5 \cite{lu2024ovis} & 63.7 & \cellcolor[HTML]{F2F3F5} 63.4 \\
     ~ & MathVerse$_{(\text{vision-only})}$ \cite{zhang2024mathverse} & VQA & 54.8 \cite{google2024gemini} & 26.6 \cite{liu2024points1} & 16.2 & \cellcolor[HTML]{F2F3F5} 19.0 \\
     ~ & MathVision$_{(\text{full})}$ \cite{wang2024measuring} & VQA & 43.6 \cite{google2024gemini} & 22.0 \cite{liu2024points1} & 17.8 & \cellcolor[HTML]{F2F3F5} 18.8 \\
    \midrule
    Text-Rich & TextVQA$_{(\text{val})}$ \cite{singh2019towards} & VQA & 82.0 \cite{megvii2024taiyi} & 78.5 \cite{li2024llavaonevisioneasyvisualtask} & 78.2 & \cellcolor[HTML]{F2F3F5} 81.3 \\
    ~ & ChartQA$_{(\text{test})}$ \cite{masry2022chartqa} & VQA & 81.2 \cite{megvii2024taiyi} & 82.4 \cite{yao2024minicpm} & 82.2 & \cellcolor[HTML]{F2F3F5} 80.5 \\
    ~ & OCRBench \cite{2024liuocr} & VQA & 89.4 \cite{sensetime2024sense} & 82.2 \cite{chen2024expanding} & 69.0 & \cellcolor[HTML]{F2F3F5} 70.0 \\
    \bottomrule
    \end{tabular}}
    \label{tab:chat}
    \vspace{-12pt}
\end{table*}

\subsection{Evaluation Results of \rewardname}
\noindent \textbf{Benchmarks.}
To evaluate \rewardname, we use diverse reward model benchmarks: \textbf{(1)} VL-RewardBench \cite{li2024vlrewardbench}, encompassing 1250 multi-modal problems addressing general understanding, hallucination, and reasoning challenges; (2) Reward-Bench \cite{lambert2024rewardbench}, with 2985 language-only problems including chat, chat hard, safety and reasoning; and \textbf{(3)} RM-Bench \cite{liu2024rm}, comprising 1237 language-only problems across chat, math, code, and safety.
RM-Bench defines three tracks (easy, normal, hard) that evaluate the sensitivity of reward models to subtle content variations and style biases.
While Reward-Bench and RM-Bench are designed for reward models of language-only LLMs, we evaluate \rewardname on these benchmarks to demonstrate that our multi-modal reward model maintains strong language capabilities despite also processing image and video inputs.

\subsubsection{Results on VL-RewardBench}
\noindent \textbf{Main Results.} Tab. \ref{tab:vl_reward} presents the evaluation results of various multi-modal RMs on the VL-RewardBench \cite{li2024vlrewardbench}.
Unlike previous multi-modal generative reward models, our \rewardname is a discriminative model that predicts a scalar reward.
Our proposed \rewardname model, despite being an open-source 7B parameter model, outperforms all other open-source models.
Notably, \rewardname achieves the highest overall accuracy (65.8\%) among open-source models and the highest Macro Accuracy (70.0\%) among all models, indicating its superior performance in handling diverse tasks within the VL-RewardBench.

\noindent \textbf{Strong Performance on General Problems.} The results in Table \ref{tab:vl_reward} reveal that \rewardname achieves a significantly higher accuracy (84.7\%) on general problems compared to other generative RMs. We found the reason is attributed to these problems presenting a considerable challenge, often leading to tied judgments in previous LVLMs, whereas \rewardname demonstrates a greater ability to make correct classifications with different scalar scores.

\subsubsection{Results on Reward Bench and RM-Bench}
\noindent \textbf{Main Results.} We argue that multi-modal RMs should preserve strong language processing abilities despite the incorporation of image and video data during training. Consequently, we evaluate the performance of multi-modal reward models, including \rewardname, on Reward Bench (Tab. \ref{tab:reward-bench}) and RM-Bench (Tab. \ref{tab:rm-bench}). The results demonstrate that \rewardname achieves considerable performance and surpasses other multi-modal models on this benchmark.

\noindent \textbf{Sensitivity to Content and Style.} Consistent with findings in \cite{liu2024rm}, \rewardname demonstrates sensitivity to subtle content variations and style biases, an issue often overlooked in multi-modal research. We believe further research is needed to enhance the robustness of multi-modal reward models.

\subsection{Evaluation Results of \chatname}

\noindent \textbf{Benchmarks.} We select four representative benchmarks for evaluating the instruction following and in-the-wild chatting abilities of LVLMs. (1) The WildVision bench \cite{lu2024wildvision} uses prompts collected from user submissions, reflecting real-world multimodal interactions. (2) MIA-bench \cite{qian2024mia} that is specially designed to evaluate instruction following. (3) MM-MT \cite{agrawal2024pixtral} which is an instruction-following benchmark for multi-modal models, exhibits a strong correlation with LMSys-Vision ELO ratings \cite{chiang2024chatbot}. (4) MM-Vet \cite{yu2023mm} that evaluate LVLMs on complex tasks such as language generation.
These datasets contain open-ended questions and the referenced answers, and evaluation is performed using an LLM-as-a-Judge \cite{zheng2023judging} approach, which involves using a judge model like GPT-4o \cite{hurst2024gpt} to predict scores.

We also report the performance on other categories, such as MMBench \cite{liu2025mmbench}, MMMU \cite{yue2024mmmu} and MMStar \cite{chen2024we} for general knowledge, MathVerse \cite{zhang2024mathverse} and MathVision \cite{wang2024measuring} for math reasoning, TextVQA \cite{singh2019towards}, ChartQA \cite{masry2022chartqa} and OCRbench \cite{2024liuocr} for text-rich document understanding.
These benchmarks utilize multiple-choice questions (MCQ) or visual question answering (VQA), where responses are limited to short keywords and evaluated based on string matching.

\noindent \textbf{Results on Instruction Following \& Chat.} Tab. \ref{tab:chat} shows that \chatname outperforms previous SOTA models across multiple benchmarks (WildVision, MIA, and MM-MT), demonstrating significant improvements in multi-modal understanding with instruction following ability and providing more comprehensive information for in-the-wild chat scenarios.

\noindent \textbf{Results on Other Categories.} On other categories (Knowledge, Reasoning, and Text-Rich), \chatname performs comparably to the supervised fine-tuned (SFT) model \sftname, demonstrating that RL training with \rewardname improves instruction following and conversational ability without sacrificing performance in these areas.

\section{Conclusion and Future Work}
We present \rewardname, a multi-modal reward model that is capable of multi-modal RL training, test-time scaling, and data cleaning.
Using \rewardname, we further trained \chatname via RLHF techniques to optimize the multi-modal user chat experience, focusing on providing detailed explanations and in-depth answers.
We believe that exploring multi-modal reward models with on-policy reinforcement learning algorithms holds significant promise for future research, such as exploring reward benchmarks and RL algorithms for video alignment.

\section{Limitations}
The limitation of our work stems from the composition of our training data, which is primarily sourced from English language corpora.
This reliance on English-centric data potentially limits the multilingual capabilities of our reward model. 
The English language datasets may reflect specific cultural viewpoints and societal biases prevalent in English-speaking communities.
Future research should consider the incorporation of multilingual datasets to mitigate these limitations and enhance the generalizability and fairness of the multi-modal reward model.

\bibliography{custom}

\clearpage
\appendix

\section*{\centering Appendix}

\section{More Experimental Results}
\label{sec:appendix}

\paragraph{Implementation Details}
For \rewardname, the learning rates were set at 1e-5 with a batch size of 256.
As for \chatname, the learning rates were set at 5e-5 with a batch size of 256.
We set the PPO hyper-parameters $\gamma=0.99$, $\beta=0.95$, and $\epsilon=0.2$.

\begin{table}[h]
    \centering
    \caption{\textbf{Ablation Studies} of the impact of response length constraints of reward models that guided training \chatname.}
    \resizebox{.48\textwidth}{!}{
    \begin{tabular}{l|c|cccc}
     ~ & Avg & Wild & \multirow{2}{*}{MIA} & \multirow{2}{*}{MM-MT} & MM-Vet \\
     ~ & Tokens & Vision & ~ & ~ & v2 \\
     \midrule
     w/o Length Constraints & \textbf{361} & \textbf{76.2} & \textbf{87.0} & \textbf{5.86} &  \textbf{56.6} \\
     \chatname & 274 & 74.6 & 84.0 & 5.70 & 54.8  \\
    \end{tabular}}
    \label{tab:ablation_length}
    \vspace{-12pt}
\end{table}

\paragraph{The Impact of Length Constraints} 
To prevent the chat model from generating overly long responses to artificially inflate rewards, we introduce length constraints on the ratio of chosen to rejected responses during training reward model \rewardname.
The ablation study results of length constraints are present in Tab. \ref{tab:ablation_length}.
On the WildVision benchmark, we compute the average token length of the model's responses. We observe a substantial increase in average token length, from 274 to 361, when length constraints were not applied.
Surprisingly, removing length constraints yielded substantial improvements in open-ended benchmarks, achieving state-of-the-art results.
Such observation is because these benchmarks do not penalize length in their evaluation prompts, judge models (e.g., GPT-4) tend to favor longer responses, even if they contain unnecessary details that detract from the user experience.
As our focus is on optimizing user experience, not benchmark scores, we retain the length constraints.
Following the precedent set by language-only benchmarks (e.g., \cite{dubois2024length}), we believe multi-modal Chat Arena and dialogue benchmarks should also address potential length and style biases in their evaluation protocols in future work.

\paragraph{Results on Test-Time Scaling}
According to Tab. \ref{tab:test_time_scaling}, we observe that the Best-of-$N$ sampling further improves the results. The averaged tokens is increased slightly (from 274 to 283), demonstrate that the improvements is bring from the high-quality response, rather than hacking the length bias in Tab. \ref{tab:ablation_length}. 

\begin{table}[h]
    \centering
    \caption{\textbf{Results of Best-of-$N$ (BoN) sampling} for test-time scaling with \rewardname.}
    \resizebox{.48\textwidth}{!}{
    \begin{tabular}{lc|c|cccc}
     ~ & \multirow{2}{*}{N} & Avg & Wild & \multirow{2}{*}{MIA} & \multirow{2}{*}{MM-MT} & MM-Vet \\
     ~ & ~ & Tokens & Vision & ~ & ~ & v2 \\
     \midrule
     \chatname & ~ &274 & 74.6 & 84.0 & 5.70 & 54.8  \\
     \chatname + BoN & 4 & 283 & \textbf{77.7} & \textbf{87.3} & \textbf{6.03} & \textbf{56.3} \\
    \end{tabular}}
    \label{tab:test_time_scaling}
    \vspace{-12pt}
\end{table}

\begin{figure*}
\centering
\includegraphics[width=0.95\linewidth]{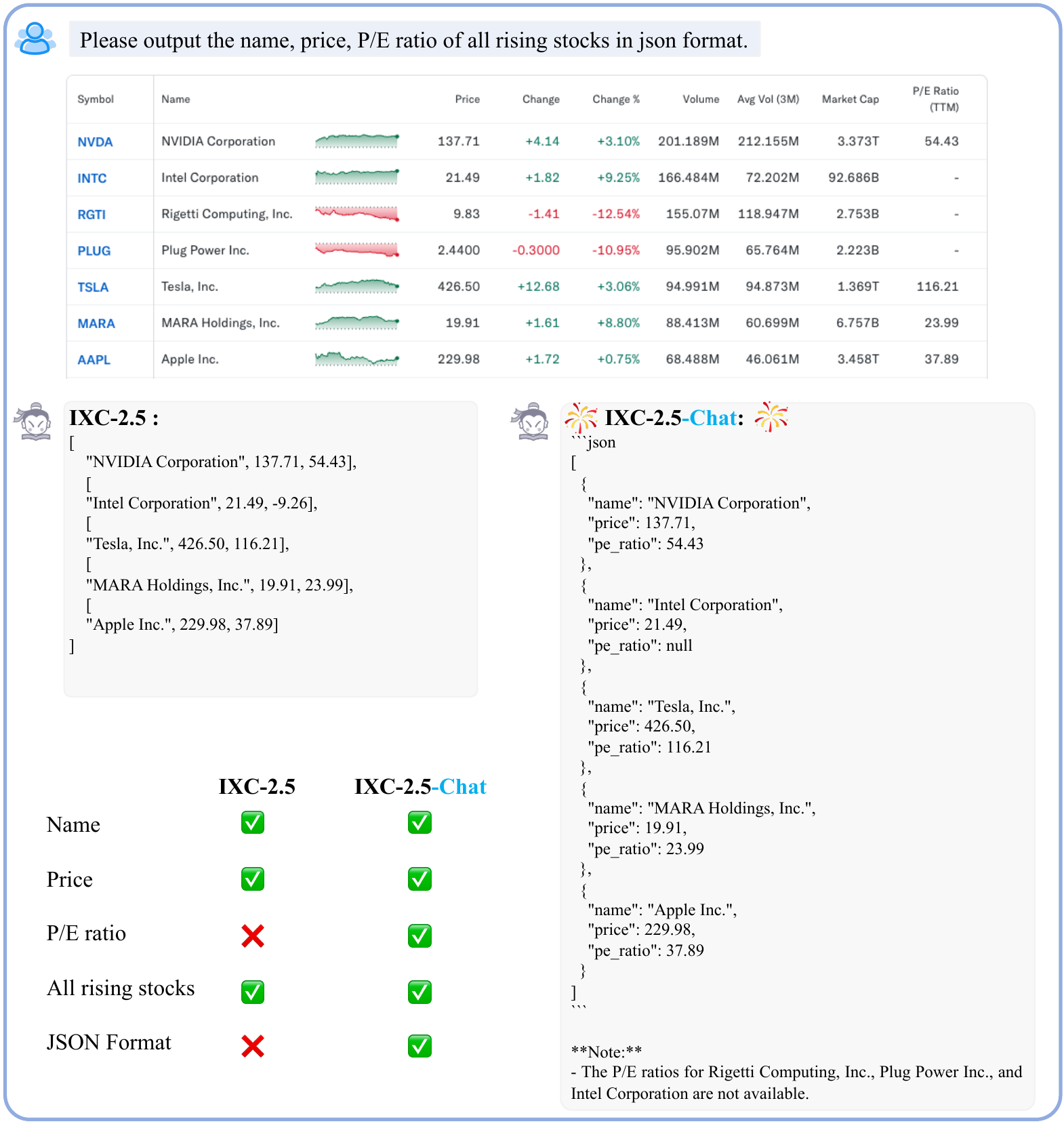}
\vspace{-6pt}
\caption{\textbf{Visualizations} of multi-modal dialogues generated by \chatname on \textbf{instruction following} abilities.}
\label{fig:visualize_if}
\end{figure*}

\begin{figure*}
\centering
\includegraphics[width=0.95\linewidth]{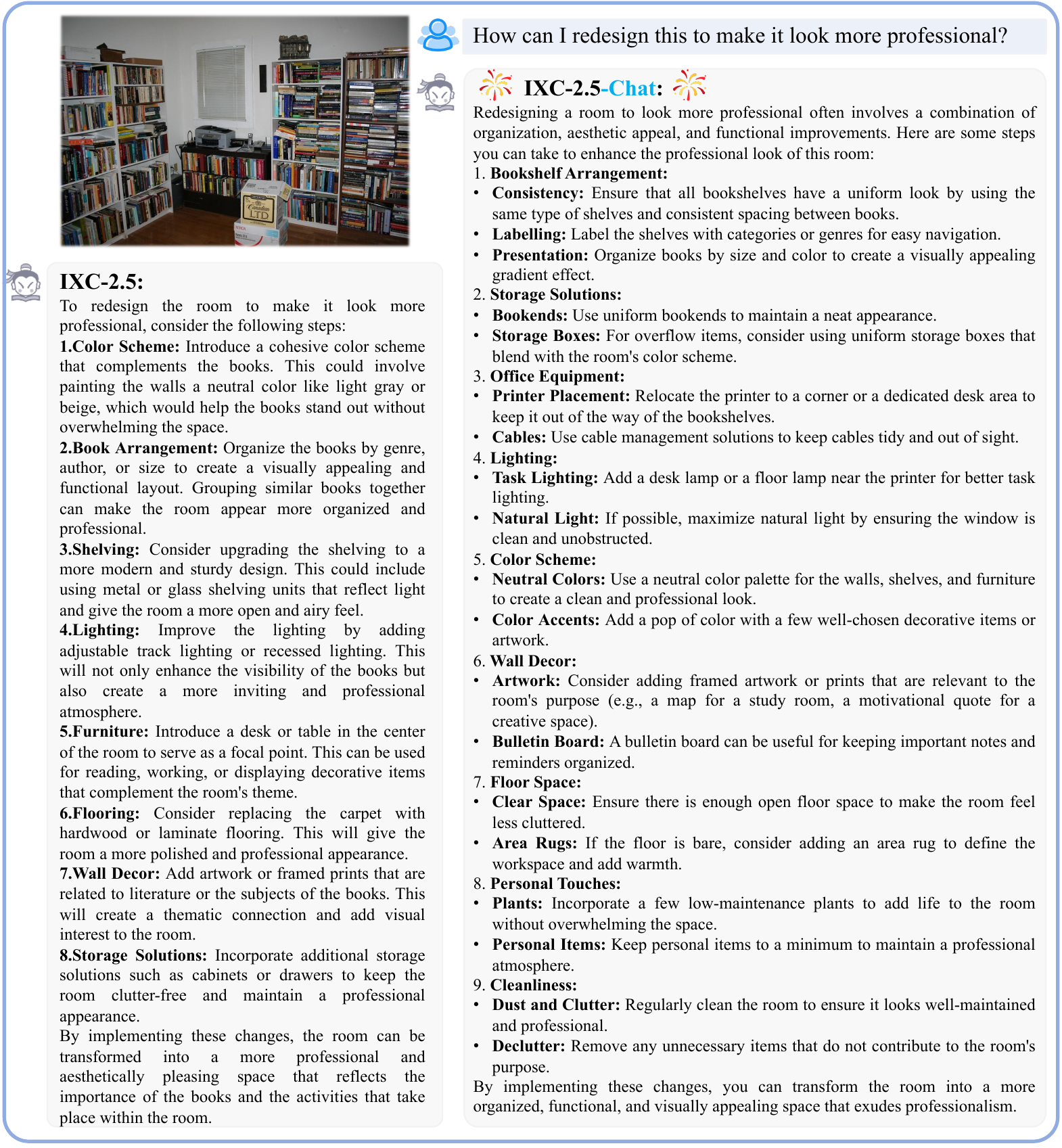}
\vspace{-6pt}
\caption{\textbf{Visualizations} of multi-modal dialogues generated by \chatname on \textbf{open-ended} questions.}
\label{fig:visualize_open}
\end{figure*}

\paragraph{Visualization Results}
We present the visualization examples of \chatname on a series of topics, such as instruction following (Fig. \ref{fig:visualize_if}) and open-ended questions (Fig. \ref{fig:visualize_open}). These figures reveal that \chatname demonstrates several key advantages, including superior organization and presentation, more comprehensive and in-depth answers, and more detailed explanations. These strengths significantly enhance \chatname's effectiveness in multi-modal chat interactions.

\end{document}